# Agreement Between Large Language Models and Human Raters in Essay Scoring: A Research Synthesis

Hongli Li*, Che Han Chen, Kevin Fan, Chiho Young-Johnson, Soyoung Lim, and Yali Feng

** Correspondence: hli24@gsu.edu. Department of Educational Policy Studies, Georgia State University*

**Abstract**

Despite the growing promise of large language models (LLMs) in automated essay scoring (AES), empirical findings regarding their reliability compared to human raters remain mixed. Following the PRISMA 2020 guidelines, we synthesized 65 published and unpublished studies from January 2022 to August 2025 that examined agreement between LLM-generated scores and human ratings. Agreement levels varied substantially both across and within studies, with reported values spanning a wide range. Overall, the findings suggest that LLM–human agreement is highly context-dependent. Implications, challenges, and directions for future research are discussed.

# Introduction

Automated essay scoring (AES) refers to automatically assigning a score to an essay in a way that approximates or complements human judgment (Dikli, 2006; Shermis & Burstein, 2013). The origins of AES can be traced back to the Project Essay Grade (PEG; Page, 1966), which used linear regression models based on surface features such as word count, sentence length, and punctuation to predict the essay scores that human raters would assign. Later systems, including *e-rater* (Attali & Burstein, 2006) and *IntelliMetric* (Rudner et al., 2006), advanced this approach by incorporating a larger set of linguistic and syntactic features and by applying machine learning algorithms. These feature-based AES models were typically developed within testing organizations and trained on human-scored essays to predict scores from measurable linguistic indicators. Such models have demonstrated good consistency with human ratings within the same datasets or writing prompt types used for training (Shermis & Burstein, 2013), but they often struggled to generalize across different writing contexts or genres (Deane, 2013; Williamson et al., 2012).

More recent transformer-based models, such as BERT (Bidirectional Encoder Representations from Transformers, Devlin et al., 2019) and DeBERTa (Decoding-enhanced BERT with disentangled attention, He et al., 2021), introduced deep contextual representations and improved performance in text-related tasks, including essay scoring. These models have also been applied to automated essay scoring and fine-tuned on datasets such as the Automated Student Assessment Prize (ASAP) dataset (Hamner et al., 2012). However, these models typically rely on supervised fine-tuning with labeled datasets and substantial technical expertise. They do not operate in the same way as human raters in essay scoring, thus limiting their use largely to researchers. Therefore, they were not included in the present synthesis.

Instead, the present synthesis focuses on generative transformer-based large language models (LLMs), such as GPT, Claude, Gemini, and LLaMA. These models are primarily developed by major technology companies (e.g., OpenAI, Anthropic, Google) and are increasingly adapted or extended by academic and open-source communities. These LLMs are trained on massive text corpora and can generate, interpret, and evaluate text through contextual understanding without task-specific training (OpenAI, 2023). Because of their capacity for semantic reasoning, flexible prompting strategies, and accessibility through public interfaces and APIs, these LLMs have rapidly become attractive tools for automated essay scoring in a variety of contexts.

Before the widespread adoption of transformer-based LLMs, Shi and Aryadoust (2024) conducted a comprehensive review of artificial intelligence-based automated written feedback (AWE) systems published between 1993 and 2022. They defined AWE as "a type of artificial intelligence (AI) that uses natural language processing algorithms to automatically assess and provide feedback on written language productions" (p. 187). Their review synthesized early systems developed primarily on machine learning and natural language processing methods (such as *e-rater* and *Criterion*), but did not include LLMs, because LLMs only became widely available after the public release of ChatGPT in late 2022.

Despite the promise of LLMs in automated essay scoring, empirical findings on their reliability compared to human raters remain mixed and inconclusive. Recently, Huang et al. (2025) conducted an early review of LLMs in automated writing evaluation based on 29 studies published between 2022 and July 2025. Their search focused heavily on studies using large, publicly available datasets, for example, 14 out of the 29 studies using the ASAP dataset and five using the TOEFL11 dataset. As a result, their review provided limited coverage of researcher-collected essays in classroom settings.

To address this gap, we synthesized published and unpublished studies from 2022 to August 2025 that reported agreement between LLM-generated scores and human ratings, including both publicly available datasets and researcher-collected classroom data. In educational settings, essays are assessed for different purposes and within different instructional contexts. For example, some studies involve essay writing in English as a foreign or second language (EFL/ESL) contexts, while others examine essay writing in students' first language (L1) for courses such as language arts or writing. There are also studies focusing on content-area courses, where essays are used to assess subject-matter knowledge and genre-specific competence. Consistent with this diversity, the studies included in this review span both L1 and second language (L2) writing contexts and include tasks from language-learning settings as well as content areas. While most studies focus on essays written in English, we also include studies reporting automated essay scoring in other languages.

In most studies, human ratings are treated as the benchmark against which the performance of AES systems, including LLMs, is evaluated. However, this assumption has its limitations because of variability and disagreement among human raters themselves. In this synthesis, we still use LLM–human agreement as a reference point, while acknowledging these limitations. Furthermore, various metrics have been used to quantify LLM–human agreement in the collected studies, with the most common being Quadratic Weighted Kappa (QWK). QWK estimates chance-corrected agreement for ordinal ratings and applies quadratic weights to penalize larger discrepancies more heavily, making it well-suited for essay scoring contexts where score distances are meaningful (e.g., Cohen, 1988; Landis & Koch, 1977). For this reason, QWK has been widely used in AES research because it accounts for both chance agreement and the magnitude of scoring discrepancies. The next most frequently reported index is Pearson's $r$, followed by Spearman's $\rho$. These are measures of association rather than agreement. Therefore, they can yield high values even when absolute agreement is low (e.g., Lin, 1989; McHugh, 2012). In this synthesis, we prioritize QWK as the primary index for coding and interpreting agreement between human and LLM essay scores.

To summarize, our goal was to provide an up-to-date summary of LLM performance in essay scoring and to identify emerging patterns across studies. Specifically, we ask two research questions: (1) What are the characteristics of the included studies on LLM-based automated essay scoring? and (2) How well do LLMs perform in automated essay scoring, as reflected in their agreement with human raters?

# Methods

This synthesis followed the Preferred Reporting Items for Systematic Reviews and Meta-Analyses (PRISMA) 2020 guidelines (Page et al., 2021) to ensure a transparent and systematic review process. PRISMA provides a structured framework for identifying, screening, and selecting studies, as well as for documenting coding and synthesis procedures. In this review, we first describe the criteria used to determine study eligibility, followed by the search and screening process, and finally the coding procedures used to extract and synthesize data across studies.

## Eligibility of Studies

Studies were included in this synthesis if they met the following criteria:

a) The study appeared between January 1, 2022, and August 31, 2025. This time frame was chosen because LLMs only became widely accessible to researchers after late 2022, following the public release of ChatGPT.
b) The study used an LLM to rate essays. For the purpose of this review, LLMs refer broadly to transformer-based generative models such as ChatGPT, Gemini, Claude, LLaMA, and similar systems.
c) The study reported at least one numeric index of agreement between LLM-generated and human-assigned scores.
d) The study was written in English, either published or unpublished studies (e.g., preprints, conference papers, or technical reports).

## Searching and Screening

As summarized in Figure 1, we conducted a comprehensive search across multiple databases, including ERIC & APA PsycINFO, Web of Science, ProQuest Central, IEEE Xplore, and arXiv, yielding 291 records. The search query used was (essay or writing) AND ("language model*" or LLM or GPT or ChatGPT or "Generative AI") AND (agreement or correlation or consistency or reliability or concordance). Then a hand search of Google Scholar and reference chasing (e.g., Huang et al., 2025) further added 18 records.

After removing 61 duplicates, we screened 248 abstracts. A total of 142 records were excluded, mostly because they did not address automated essay scoring. The remaining 106 full-text reports were further assessed for eligibility. Among these, 41 were excluded, mostly because of a lack of quantitative agreement data between AI- and human-generated scores.

Ultimately, 65 studies met the inclusion criteria and were retained for synthesis (see Appendix A for the list of studies). All screening and inclusion decisions were made by multiple coders, with any discrepancies resolved through discussion and consensus to ensure consistency in the review process.

## Coding Procedures

Based on the literature (e.g., Huang et al., 2025; Shi & Aryadoust, 2024), the six authors co-developed a detailed coding form to extract key information from each study. The coding form includes two parts: study characteristics and the LLM–human agreement indices. Each researcher pilot-tested the coding form with several studies, followed by two calibration meetings to refine the coding scheme. For example, we initially included two variables: whether prompt engineering was used and whether the model was fine-tuned. However, these variables were ultimately dropped because the information was often unclear, difficult to infer, or missing across studies. Although we did not calculate formal inter-coder reliability indices, we established agreement through trial coding and iterative discussions. Subsequently, five coders independently coded their assigned subsets of studies. The first author then double-coded all studies, and any discrepancies were resolved through team discussion until consensus was reached.

Coding the agreement indices was the most challenging part, because studies reported agreement in highly diverse ways. First, different agreement metrics were used across studies, mostly including QWK, Pearson's $r$, and Spearman's $\rho$, making direct comparisons difficult. Second, many studies reported multiple (even dozens of) estimates per model or AI prompt variant. Some studies presented only the best agreement, while others reported average values or a range of agreement levels. These inconsistencies in both metrics and reporting practices made it difficult to synthesize agreement values quantitatively across studies.

To enhance comparability of agreement levels across studies, we applied a hierarchical coding rule when multiple agreement indices were reported. Specifically, we prioritized QWK, followed by Pearson's $r$, and then Spearman's $\rho$. For example, if a study reported QWK along with other indices, only QWK was coded. If a study reported both Pearson's $r$ and Spearman's $\rho$, only Pearson's $r$ was coded. If Spearman's $\rho$ was reported along with any additional indices, Spearman's $\rho$ was reported. Furthermore, when a single overall agreement estimate was reported in the study, we recorded this value. When multiple estimates were provided (e.g., across AI prompts, traits, models, rating tasks, or rating occasions), we recorded the range. Given the heterogeneity in agreement metrics and reporting practices, we did not compute pooled mean or median estimates. Instead, we reported either an average provided by the authors or the observed range.

**Figure x.1**

*Study Search Procedures*

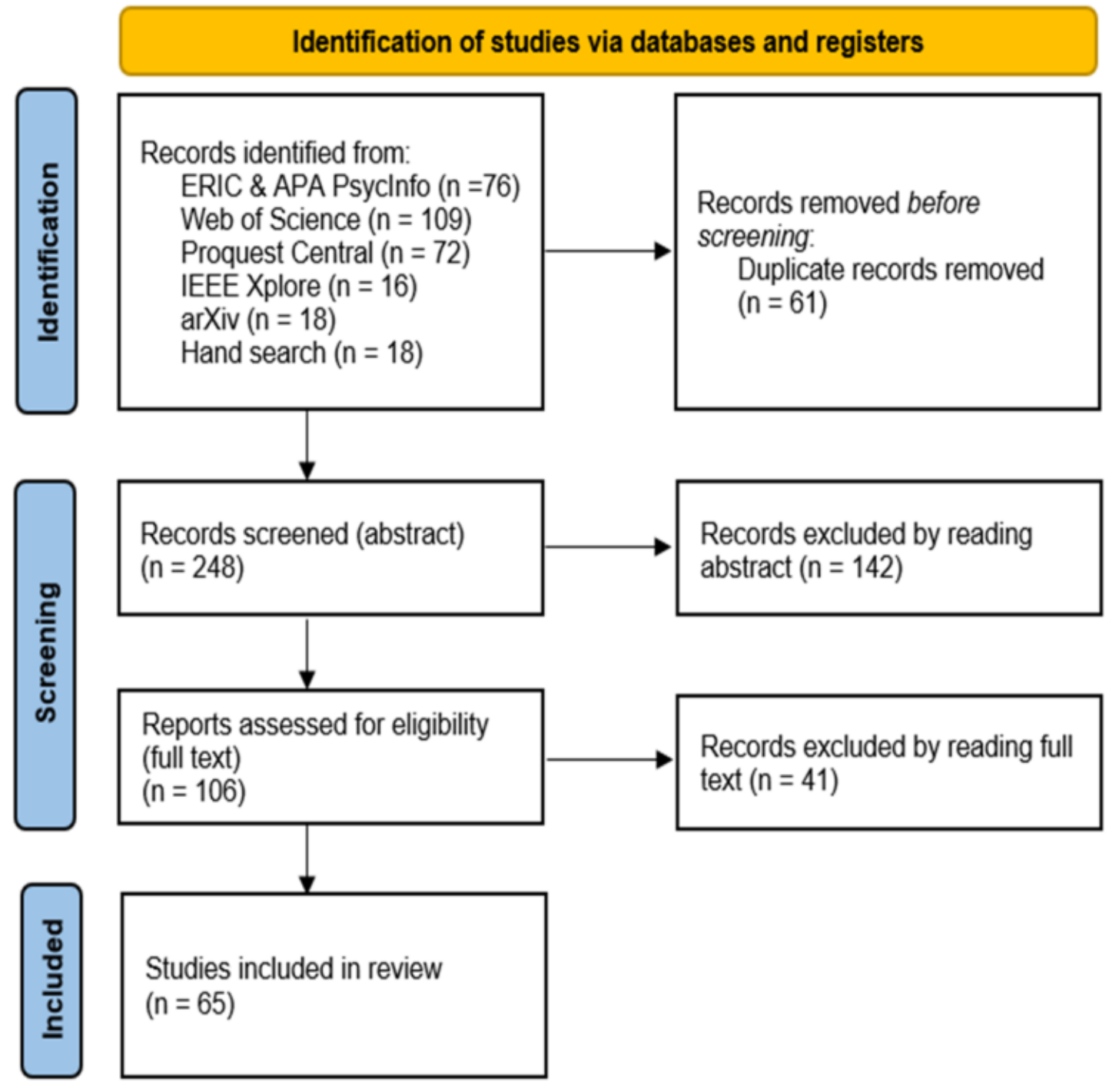

# Results

## Research Question 1: What are the Characteristics of the Included Studies on LLM-based Automated Essay Scoring?

### Study-level Characteristics (Time Period, Study Type, Country, Discipline)

As shown in Table 1, of the 65 included studies, the number of studies increased over time, with only three in 2023, 31 in 2024, and 31 in 2025 (as of August 31). The sharp increase in studies during 2024 and 2025 reflects growing research interest in LLM-based essay scoring. It is expected that more studies will appear in the coming years as AI technology continues to evolve and gain wider adoption.

Study types were diverse. Among the 65 studies, most were journal articles ($n$ = 35), followed by arXiv preprints ($n$ = 17), conference papers ($n$ = 12), and book chapters ($n$ = 1). This distribution suggests that research on LLM-based essay scoring is fast-paced and expanding rapidly, with many findings still in preprint or conference paper stages.

**Table x.1**

*Study-Level Characteristics*

| Characteristics | Subcategory | Frequency |
|---|---|---|
| Time period | 2023 | 3 |
| | 2024 | 31 |
| | 2025 as of August 31 | 31 |
| Study type | Journal article | 35 |
| | ArXiv preprints | 17 |
| | Conference paper | 12 |
| | Book chapter | 1 |
| Country of first author affiliation | China | 19 |
| | the United States | 12 |
| | Korea | 5 |
| | Germany | 4 |
| | Turkey | 4 |
| | Japan | 3 |
| | Australia | 2 |
| | India | 2 |
| | Vietnam | 2 |
| | Other countries | 13 |
| Discipline of the first author | Linguistics / Applied Linguistics / TESOL | 19 |
| | Computer Science / Engineering / Data Science | 18 |
| | Education / Educational Technology / Psychology | 13 |
| | Other fields | 13 |
| | Unknown | 2 |

Among the 65 studies, most first authors were affiliated with institutions in China ($n$ = 19) and the United States ($n$ = 12), followed by Korea ($n$ = 5) and Germany ($n$ = 4). Smaller numbers were from Turkey ($n$ = 4), Japan ($n$ = 3), Australia ($n$ = 2), India ($n$ = 2), and Vietnam ($n$ = 2). This distribution indicates that research on LLM-based essay scoring is highly international, with strong representation from China and the United States. It is noteworthy that Gjorevski et al. (2025) was counted for both the US and Germany, because the first author has two affiliations in these two countries.

The distribution of first authors' academic fields reflects the strong interdisciplinary nature of research on LLM-based essay scoring. Most studies were led by scholars in linguistics and applied linguistics ($n$ = 19) and computer science or data science ($n$ = 18), followed by those in education and educational technology ($n$ = 13). Other fields include economics, management, informatics, dentistry, social work, etc. This shows the active involvement of both content experts, who bring theoretical and linguistic insights into writing assessment, and method specialists, who contribute computational and algorithmic expertise, in LLM-based essay scoring.

## Characteristics of Participants and Writing Contexts

This section summarizes key characteristics of participants and writing contexts in the included studies (see Table 2). Specifically, we examined participants' education level, whether the essay language was the students' L1 or L2, the language of the essays, the type of scoring rubric used (analytic vs. holistic), and the types of writing tasks (i.e., language-learning vs. content-area tasks).

Participants across studies varied widely in age and educational level. However, not all studies reported detailed participant information, and some included mixed-age samples. Among the 50 studies that clearly reported such information, the majority involved undergraduates ($n$ = 24) or secondary students ($n$ = 19), while four studies included postgraduates or adults. Only three studies focused on elementary students, for example, Song et al. (2024) with third graders, Feng et al. (2024) with primary school students from K to 6, and Nakamoto et al. (2024) with fifth graders. This limited representation of younger learners indicates an important research gap.

**Table x.2**

*Characteristics of Participants and Writing Contexts*

| Characteristics | Subcategory | Frequency |
|---|---|---|
| Education Level | Undergraduate | 24 |
| | Secondary | 19 |
| | Postgraduate/Adult | 4 |
| | Elementary | 3 |
| | Unknown or mixed | 15 |
| Whether the essay language is students' L1 or L2 | L2 | 31 |
| | L1 | 22 |
| | Unknown | 12 |
| Language of the essay | English | 53 |
| | Chinese | 6 |
| | Japanese | 2 |
| | Arabic | 1 |
| | German | 1 |
| | Russian | 1 |
| | Turkish | 1 |

**Table x.2 (continued)**

*Characteristics of Participants and Writing Contexts*

| Characteristics | Subcategory | Frequency |
|---|---|---|
| Rubrics | Analytic | 37 |
| | Holistic | 26 |
| | Both | 1 |
| | Unknown | 1 |
| Writing tasks | Language learning | 55 |
| | Content learning | 7 |
| | Unknown or mixed | 3 |

In many cases, the essay language was not necessarily the students' L1. For example, in Koraishi (2024), participants were Turkish secondary students who responded to IELTS writing tasks, with English as their L2. Xiao et al. (2025) involved Chinese junior high school students writing narrative essays in Chinese, where the essay language represented their L1. Among the studies that clearly reported this information, 31 studies involved L2 writers, and 22 studies included L1 writers. This pattern shows that LLM-based essay scoring has been more heavily investigated in second-language writing contexts. It is important to note that the L1/L2 classification and writing task categories represent different dimensions and are not directly comparable. For example, studies using the ASAP dataset involve essays written by U.S. students in their L1 but are still considered language learning tasks.

Essays were predominantly in English (*n* = 53) with some in Chinese (*n* = 6), Japanese (*n* = 2), Arabic (*n* = 1), German (*n* = 1), Russian (*n* = 1), and Turkish (*n* = 1). This indicates that most existing studies on LLM-based essay scoring have focused on English-language writing, reflecting both the dominance of English in academic communication and the fact that most LLMs are primarily trained on English text. Therefore, much less is known about how well these models perform in non-English contexts.

Regarding scoring methods, both analytic rubrics (*n* = 37) and holistic rubrics (*n* = 26) were used. For instance, Uyar and Büyükahiska (2025) employed the IELTS Task 2 analytic rubric, while Pack et al. (2024) used a holistic rating scale, in which strong writing is defined by clear organization, the use of cohesive devices, and fully developed paragraphs with concrete supporting details. The predominance of analytic rubrics suggests that more researchers sought to examine different constructs of written performance, rather than overall text quality judgments.

Finally, students completed a variety of writing tasks. We broadly classified the writing tasks into two categories. The first category included essays written in language-learning contexts. Fifty-five studies fell into this category. The second category comprised essays written in content-area courses, where the goal was to evaluate students' subject-matter knowledge.

Seven studies were classified in this group. For example, in Li et al. (2024), undergraduate students practiced writing college English test band 4 essays, while Xu et al. (2025) involved secondary students writing physics essays. In the latter case, ratings not only considered writing proficiency but also students' content knowledge.

## Research Question 2: How Well do LLMs Perform in Automated Essay Scoring, as Reflected in Their Agreement with Human Raters?

### LLMs Used for Rating

Some studies used multiple LLMs for comparison purposes. In this analysis, each model was counted once when it was used in a study. GPT-4 was the most frequently used model ($n$ = 32), followed by GPT-3.5 ($n$ = 22). Other models included LLaMA, Mistral, and Qwen (16 combined), Claude ($n$ = 5), and Gemini ($n$ = 3). Additional models used less frequently included PaLM (Pack et al., 2024; Liew & Tan, 2024), Bard (Masikisiki et al., 2023), Aya, Jais, and ACEGPT (Ghazawi & Simpson, 2025), Baichuan-13B and InternLM-7B (Song et al., 2024), as well as DeepSeek and several other models (see Su, Yan, Fu, et al., 2025 for a list of models).

Overall, the distribution indicates that, thus far, research on automated essay scoring has been dominated by OpenAI's GPT models, likely due to their early accessibility and strong performance in language capabilities. However, the increasing appearance of other open-source models shows a trend of using more diversified LLMs for this task.

#### LLM–Human Agreement Levels

Among the 65 studies, we coded QWK values for 38 studies and coded Pearson's $r$ or Spearman's $\rho$ for 22 studies. Five studies were not coded because they used other indices, such as intraclass correlation coefficient (*ICC*; Lu et al., 2024; Koraishi, 2024; Shabara et al., 2024), generalizability theory coefficients (Li et al., 2024), and root mean square error (*RMSE*; Nakamoto et al., 2024).

Following the guidelines of Altman (1991), adapted from Landis and Koch (1977), QWK values below 0.20 indicate poor agreement; 0.21–0.40, fair; 0.41–0.60, moderate; 0.61–0.80, good; and 0.81–1.00, very good agreement. According to Cohen (1988), correlation coefficient values around 0.30 and 0.50 are considered medium and large effects. However, correlation coefficients reflect association rather than exact agreement and do not have directly comparable benchmarks for agreement interpretation. Due to inconsistent reporting practices across studies, we did not compute pooled mean or median estimates. Instead, we relied on graphic summaries and descriptive comparisons to characterize the distribution of agreement values.

We first examined the distribution of QWK values in Figure 2. QWK values varied widely, ranging from 0.00 (Kim, 2025) to 0.97 (Oketch et al., 2025). Except for a small number of studies falling into very low (such as $< 0.01$) or very high ($> 0.9$) agreement categories, the majority of the studies reported QWK values larger than 0.30. Overall, the plot shows that QWK values are distributed across a broad range rather than concentrated at a single level.

We then examined the distribution of the reported Pearson's *r* and Spearman's ρ values (Figure 3). Reported values varied widely, ranging from as low as 0.05 (Masikisiki et al., 2023) to as high as 0.91 (Xu et al., 2025). Compared to QWK, correlation-based indices showed slightly stronger values, with many falling between 0.50 and 0.80. This difference is expected, as correlation indices measure association rather than exact agreement and do not account for chance agreement as QWK does.

In addition to variation across studies, variability was also observed within studies. Many studies reported multiple agreement values across different prompts, scoring traits, or model configurations. In some cases, the range of values within a single study was comparable to the variation observed across studies. For example, in Lee et al. (2024), QWK values ranged from 0.03 to 0.59 across combinations of LLMs, prompting strategies, and datasets. Similarly, in Masikisiki et al. (2023), Pearson's *r* values ranged from 0.05 to 0.83 across different LLMs. These examples illustrate that agreement between LLM-generated and human ratings could vary greatly even within studies.

We also observed some descriptive tendencies across studies. For instance, GPT-4 usually outperformed GPT-3.5, likely due to advances in model capability. Stronger agreement was also observed when models were guided by a detailed scoring rubric and supported with writing examples (e.g., Masikisiki et al., 2023; Yancey et al., 2023). Studies using standardized datasets such as ASAP or TOEFL11 tended to report higher agreement values (e.g., Lee et al., 2024; Wang & Gayed, 2024) compared with studies based on smaller, classroom writing tasks. However, these patterns are based on descriptive observations rather than systematic analysis.

**Figure x.2**

*Distribution of QWK Values*

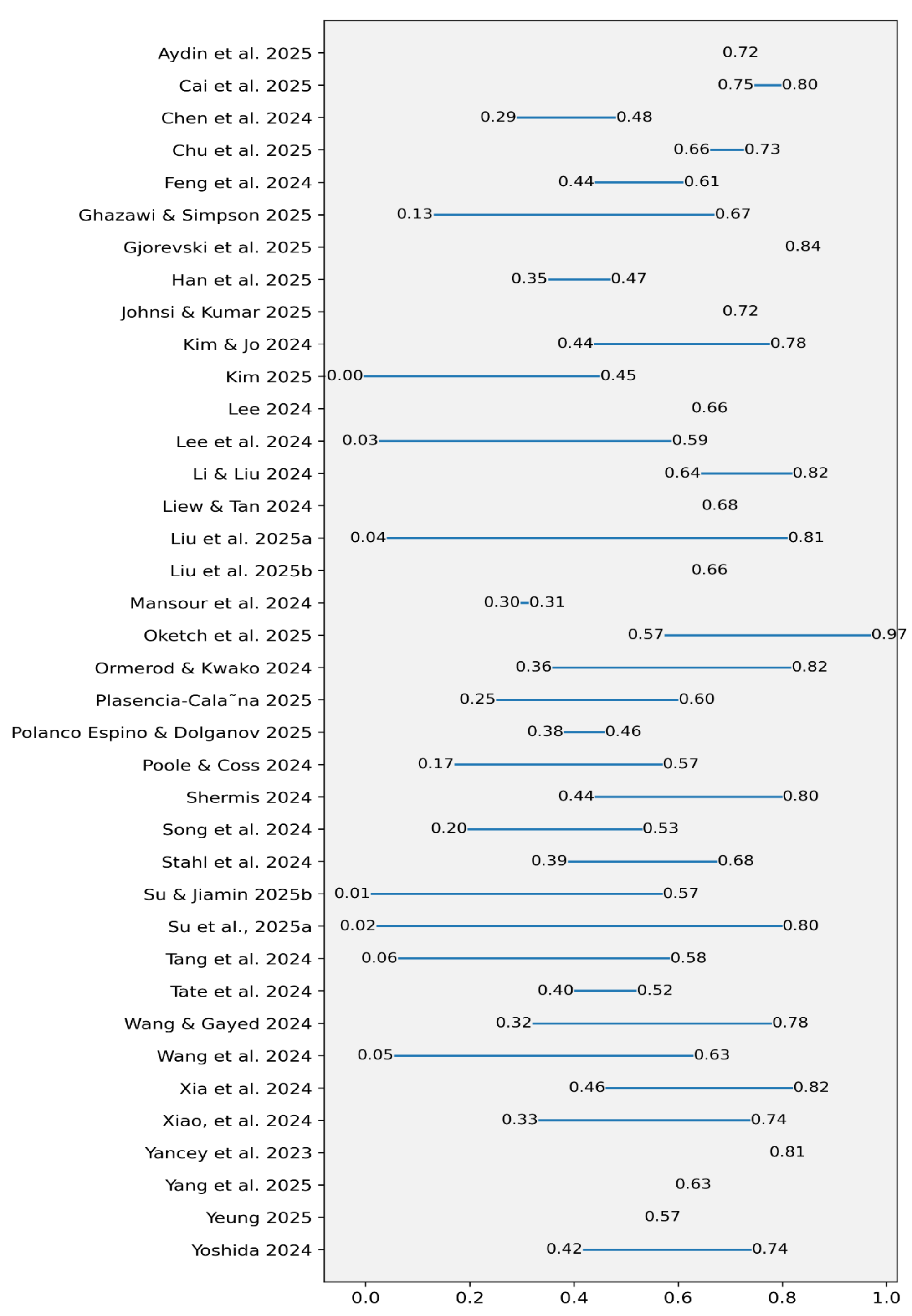

**Figure x.3**

*Distribution of Pearson's r and Spearman's ρ*

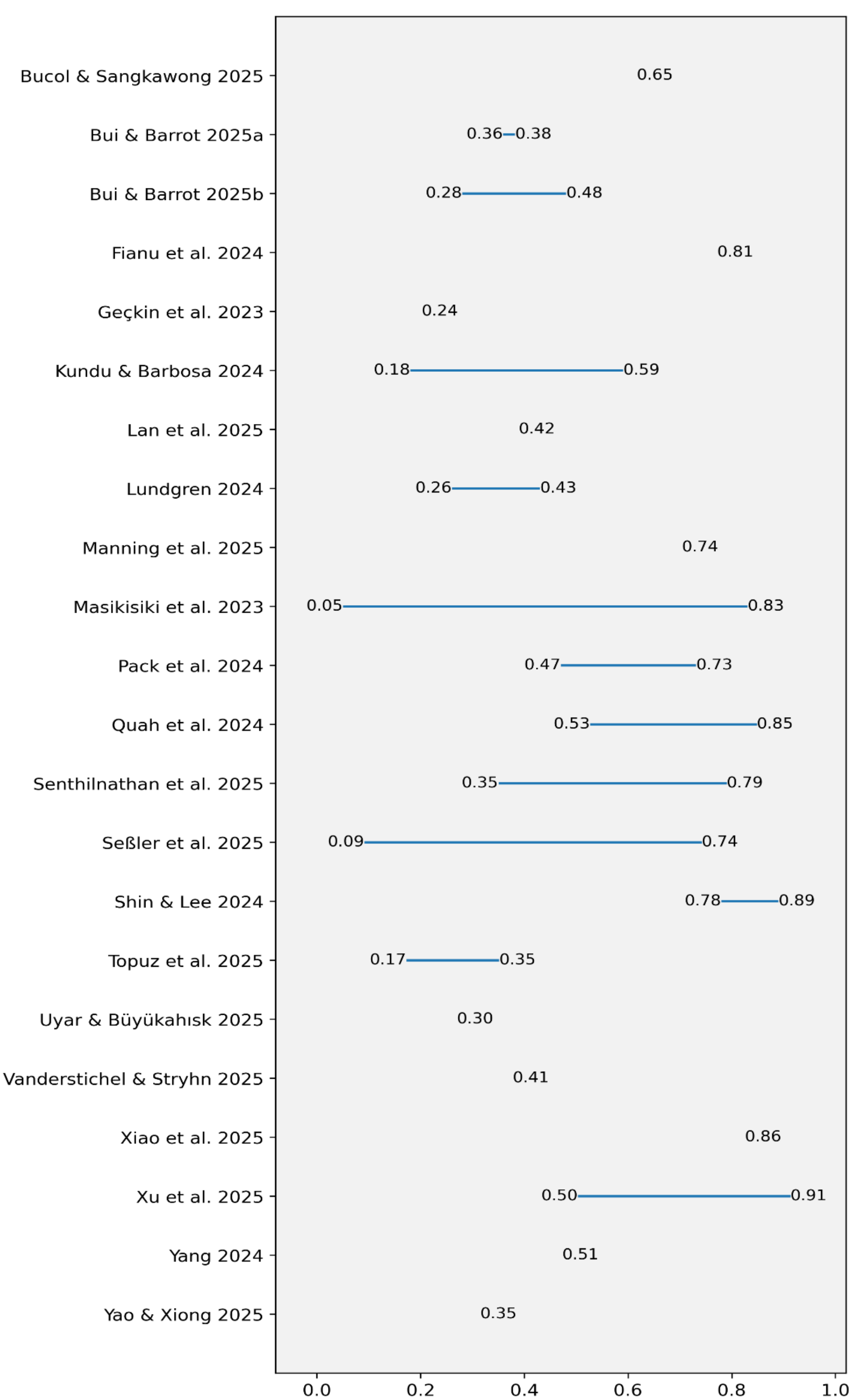

## Discussion

We found that agreement between LLM-generated scores and human ratings varied substantially across studies, with reported values spanning a wide range. This raises an important question about whether the observed levels of agreement are sufficient to support the use of LLMs for automated essay scoring. Based on the current evidence, such a conclusion cannot be made. According to Ramineni et al. (2012), the preferred Quadratic Weighted Kappa value between automated and human scoring is 0.70. Using this benchmark, many of the reported agreement values in the collected studies fall below this level. At the same time, some studies report higher agreement under certain conditions, indicating that LLMs can perform well in specific contexts. Overall, these findings suggest that while LLM-based scoring shows promise, its performance is not consistent across conditions. This general pattern aligns with Huang et al. (2025), who concluded that general language models "hold promise to enhance [automatic writing evaluation]" (p. 1).

Several factors may contribute to this variability observed in this synthesis. First, the studies included in this synthesis are highly diverse. To provide a comprehensive overview of the literature, we included studies across a wide range of contexts, including both published and unpublished work, language-learning and content-based writing tasks, and participants ranging from elementary students to undergraduates. The essays themselves were written in multiple languages, including English and several other languages. The heterogeneous study contexts may account for the observed substantial variation in LLM–human agreement levels.

Second, agreement may vary depending on how LLMs are implemented in each study. The models used differ in capability given that LLMs advance quickly. In addition, studies vary in their use of prompting strategies, parameter settings, and model configurations. Some studies experiment with multiple prompts and fine-tuning procedures, while others incorporate detailed scoring rubrics or example essays. Differences in scoring approaches, such as analytic versus holistic rubrics, may further influence agreement. These variations in implementation can lead to different levels of agreement between LLM and human ratings, but inconsistent reporting of these aspects across studies makes the findings difficult to compare.

Third, variability may arise from differences in how agreement is calculated and reported. Studies use different indices (e.g., QWK, Pearson's $r$, Spearman's ρ), which are not directly comparable. In addition, reporting practices vary: some studies report only the highest agreement value, whereas others present averages or full ranges across multiple runs or conditions. As a result, reported agreement levels may reflect different underlying practices. This variation may also be influenced by disciplinary differences. For example, studies in computer science often emphasize model optimization and may report best-case performance, whereas studies in language or education fields are more likely to report more comprehensive results.

Finally, variability in agreement may also reflect limitations in the reference standard itself. A low agreement between humans and LLMs could be due to limited model performance or poor human ratings. In most studies, human ratings are treated as the benchmark for evaluating LLM performance. However, prior research has shown that human raters do not always agree with one another. As a result, lower agreement between LLMs and human raters may reflect not only limitations of the models but also variability in human scoring.

We observed several descriptive tendencies, such as stronger performance of more advanced models, higher agreement with detailed rubrics and example essays, and higher agreement in studies using standardized datasets; however, these patterns are based on descriptive observations rather than systematic analysis. Because study conditions were not consistently defined, reported, or experimentally manipulated, no firm conclusions can be drawn about the effects of these factors on agreement in the current synthesis. These patterns, therefore, need more systematic investigation in future research.

# Conclusion

This review synthesized studies published between 2022 and August 2025 that compared LLMs and human raters in automated essay scoring. Overall, agreement between LLM-generated scores and human ratings varied substantially across studies. Given the substantial heterogeneity observed across studies, future research should examine agreement levels under more controlled conditions and, where feasible, through meta-analytic approaches. For example, one possible direction for a future meta-analysis is to focus on studies using standardized datasets (e.g., ASAP or TOEFL11) to reduce variability associated with task type, participant characteristics, and scoring contexts. Within such a subset of studies, it may be possible to more systematically examine the effects of factors such as prompting strategies, model configurations, and fine-tuning.

The following are a few areas that need attention. First, as a field, we need more standardized reporting practices in studies of AES. Inconsistent reporting of agreement indices, selective presentation of results, and limited transparency regarding prompting and model configurations make it difficult to compare findings across studies. Establishing clearer reporting guidelines would facilitate more rigorous synthesis and evaluation of LLM-based scoring systems in future research. Second, as LLM-based automated essay rating tools have great potential to be used in educational settings, especially K–12 classrooms (Kasneci et al., 2023), further research is especially needed to examine how automated essay scoring performs with younger writers, whose writing skills, vocabulary, and cognitive development differ substantially from those of older students (Kim et al., 2021). Moreover, because most LLMs are trained primarily on English text, future studies should also examine automated essay scoring in other languages to ensure that AI technology can benefit diverse linguistic and learner populations (Joshi et al., 2020).

# Appendix A

## Studies Included in the Synthesis

## About the Author

**Hongli Li** is a professor in the Program of Research, Measurement, and Statistics (RMS) at the Department of Educational Policy Studies, Georgia State University. Her primary research areas are applied measurement, language assessment, and quantitative methods. Her most recent work includes peer assessment and the use of AI in test development and validation.
https://orcid.org/0000-0002-1039-7270

**Che-Han Chen** holds an MA in TESOL from Ohio State University and is now a Ph.D. student of Applied Linguistics at Georgia State University. His research interests include language testing, corpus linguistics, L2 writing, computer-assisted language learning, and English for Academic Purposes.

**Kevin Fan** is a student researcher interested in the use of artificial intelligence in education and machine learning for image recognition. He currently attends Johns Creek High School and is also enrolled at Georgia Institute of Technology through a dual enrollment program.

**Chiho Young-Johnson** holds an MA in TESOL from the University of Reading and is now a Ph.D. student of Applied Linguistics & ESL at Georgia State University. Her research interests include second language assessment, second language acquisition, and corpus research.

**Soyoung Lim** is a Ph.D. student of Applied Linguistics & ESL department at Georgia State University. She serves as a graduate fellow at the Center for Research on the Challenges of Acquiring Languages and Literacy (RCALL). Her research interests lie in integrated assessment and using process data (e.g., eye-tracking, keystroke logging) to understand L2 learners' cognitive processes in writing and integrated assessments.

**Yali Feng** is completing her Ph.D. degree in the Applied Linguistics & ESL program at Georgia State University and will join the University of Mississippi as an assistant professor this summer. Her research interests include language assessment, interlanguage pragmatics, and Chinese as a second language acquisition.

**Funding**

No external funding was received for this study.

**Competing Interests**
The authors declare no competing interests.